\documentclass[sigconf]{acmart}
\usepackage{amsmath}
\usepackage{booktabs}
\usepackage{tabularx}
\usepackage{microtype}
\usepackage{tikz}
\usepackage[table]{xcolor}
\usetikzlibrary{arrows.meta}
\usepackage{enumitem}
\usepackage{booktabs}
\usepackage{multirow}
\usepackage{colortbl}
\definecolor{lightcyan}{HTML}{E0F7FA}

\setcopyright{none}
\acmConference[ ]{ }{ }{ }

\usepackage{xspace}
\newcommand{\method}{\textsc{EvalAlloc}\xspace}

\newcommand{\searchset}{\mathcal D_{\mathrm{s}}}
\newcommand{\archive}{\mathcal A}

\newcommand{\draftnote}[1]{}

\usepackage{booktabs,multirow,makecell,adjustbox,threeparttable,wasysym,pifont,appendix}

\usepackage[most]{tcolorbox}

\definecolor{takeawayblue}{RGB}{63, 101, 145}

\newtcolorbox{takeawaybox}[1]{
  enhanced,
  breakable,
  frame hidden,
  colback=takeawayblue!5,
  borderline west={1.4pt}{0pt}{takeawayblue!85!black},
  sharp corners,
  left=6pt,
  right=5pt,
  top=4pt,
  bottom=4pt,
  before skip=6pt,
  after skip=6pt,
  fontupper=\small,
  before upper={\textbf{\textit{#1.}}\enspace}
}

\renewcommand\footnotetextcopyrightpermission[1]{}

\acmISBN{}
\acmDOI{}
\setcopyright{none}

\title{Where Should the Evaluation Budget Go in LLM-Driven Agent Scaffold Evolution?}
\author{Sichun Luo$^1$, Yi Huang$^2$, Mukai Li$^1$, Haochen Luo$^1$, Guanzhi Deng$^3$, Haibo Wang$^4$\\ Jian Xu$^5$, Zefa Hu$^2$, Junlan Feng$^2$, Qi Liu$^1$\\
$^1$The University of Hong Kong \quad
$^2$JIUTIAN Research, China Mobile \\$^3$City University of Hong Kong \quad $^4$Carnegie Mellon University \quad $^5$Tsinghua University
}

\begin{document}

\begin{abstract}
Large language model (LLM)-driven agent scaffold evolution improves agent workflows through iterative candidate generation and evaluation.
Evaluation both guides subsequent search and determines the final scaffold, but evaluating every candidate on the full task set incurs substantial execution costs.
We investigate how evaluation budgets can be allocated across these two roles.
Our experiments show that sparse feedback can sustain productive evolution, while additional targeted evaluation improves final selection.
Replay analyses further show that shared informative items and a common scoring reference support comparisons among closely matched candidates.
Motivated by these findings, we propose \textbf{\method}, an evaluation allocation framework that combines sparse search-time feedback with targeted final confirmation.
A short fully evaluated warm-up establishes a shared informative core and calibrates a response model whose item parameters remain fixed during search.
After search, \method{} refits the model and applies paired successive halving, reusing accumulated observations and executing only missing evaluations.
Across four benchmarks and two evolution frameworks, \method{} achieves higher mean  performance than full evaluation in most settings while using approximately half as many evaluation cells.
\end{abstract}

\maketitle

\section{Introduction}

A scaffold for a large language model (LLM) agent is an
executable program that defines how an agent coordinates
LLM calls, maintains intermediate state, and interacts
with tools and external information
sources~\cite{yao2023react,hu2024adas,qin2024toolllm}.
For Web-facing agents, scaffolds govern how agents acquire
and use information through Web searches, document
retrieval, and interactions with online
services~\cite{nakano2021webgpt,zhou2024webarena}.
While these scaffolds have traditionally been designed
and refined manually, recent advances have enabled
their automated optimization
~\cite{hu2024adas,zhang2025aflow,gepa}.
For example, ShinkaEvolve~\cite{shinka} iteratively
proposes scaffold modifications and uses evaluation
feedback to guide subsequent evolution.

However, evaluating candidate scaffolds can incur
substantial execution costs. A single task instance
may involve multiple LLM calls, retrieval steps,
and tool interactions~\cite{mialon2024gaia,yao2024taubench},
making full evaluation across many candidates expensive.
Although parallel evaluation and asynchronous program
generation can overlap computation~\cite{funsearch},
feedback-dependent search decisions still require
evaluation results.
Existing approaches reduce evaluation costs through
sparse score estimation and evaluation
cascades~\cite{collab,simple}, treating evaluation
primarily as a fitness-estimation problem.
In scaffold evolution, however, evaluation supports
two distinct decisions: it provides fitness feedback
that shapes archive updates and subsequent
proposals~\cite{zhang2026darwin}, and it determines
which scaffold is ultimately returned.
These decisions impose different requirements:
search benefits from inexpensive yet comparable
feedback, whereas final selection requires reliable
discrimination among competitive candidates.
This raises a central question:
\emph{How should a limited evaluation budget be
allocated across these two decisions to improve
the quality of the ultimately selected scaffold?}

To answer this question, we investigate how evaluation
allocation affects evolutionary search and final
scaffold selection. Our investigation yields three
findings.
First, sparse evaluation feedback can sustain
productive evolution, whereas removing feedback
substantially degrades search quality.
Second, shared informative items improve comparisons
between closely matched candidates, while maintaining
a common scoring reference helps preserve comparability
across evolutionary iterations.
Third, search improvements are concentrated in early
iterations, whereas final selection remains sensitive
to uncertainty among top candidates.
These findings suggest that evaluation
should provide inexpensive yet comparable feedback
during search, while reserving additional observations
for final selection.

Motivated by these findings, we propose \textbf{\method},
an evaluation allocation framework that separates
sparse search-time feedback from targeted final
selection. During search, a short fully evaluated
warm-up identifies a shared informative core and
calibrates a response model. Subsequent candidates
are evaluated on sparse item sets centered on this
core, while the model's item parameters remain fixed
to maintain a common scoring reference for archive
updates and parent selection.
After search, \method{} applies warm-started paired
successive halving (SH), reusing accumulated
observations and performing only missing evaluations.
A refitted response model further supports
comparisons during later selection rounds.
We evaluate \method{} on four benchmarks and two
evolution frameworks, with three independent runs
per setting. Compared with full evaluation, \method{}
uses approximately half as many evaluation cells
while achieving competitive or higher mean
performance across the evaluated settings.
Ablation studies further show that informative
item allocation and final selection are the
main contributors to its effectiveness.
These results demonstrate the benefits of separating
sparse search-time feedback from targeted final
evaluation.

Our main contributions are summarized as follows:
\begin{itemize}[leftmargin=13pt]
    \item We investigate evaluation allocation as a decision-dependent problem in LLM-driven scaffold evolution, demonstrating that search-time feedback and final scaffold selection impose distinct requirements on evaluation resources.

    \item We propose \method{}, an evaluation
    allocation framework that combines a shared
    informative core, a fixed search-time response
    model, and warm-started paired successive
    halving to allocate evaluation effort across
    evolutionary search and final selection.

    \item We evaluate \method{} across four
    benchmarks and two evolution frameworks,
    complemented by controlled replay analyses,
    demonstrating competitive final performance
    with substantially fewer evaluation cells.
\end{itemize}

\section{Related Work}

\textbf{LLM-Driven Agent and Program Evolution.}
Recent work has explored the automated design and
optimization of agent scaffolds and executable
workflows using LLMs \cite{novikov2025alphaevolve}.
ADAS~\cite{hu2024adas} uses a meta-agent to generate
agent programs based on previous discoveries, while
AFlow~\cite{zhang2025aflow} employs Monte Carlo Tree
Search to optimize code-represented workflows.
Other approaches, including
ShinkaEvolve~\cite{shinka} and
AdaEvolve~\cite{adaevolve}, iteratively generate,
evaluate, and refine programs using evaluation
feedback to guide subsequent search.
These approaches primarily focus on improving
candidate generation and evolutionary optimization.
Our work instead examines the evaluation process
underlying this feedback-driven search, studying how
limited evaluation resources can support both
evolutionary decisions and final scaffold selection.

\textbf{Efficient LLM Evaluation.}
Reducing the cost of evaluating LLMs has received
growing attention.
Methods such as tinyBenchmarks~\cite{tiny}
use carefully selected examples and statistical
estimation to approximate full-benchmark performance,
while SparseEval~\cite{zhang2026sparseeval} optimizes
representative items for efficient score estimation
and ranking.
CollabEval~\cite{collab} exploits shared information
across evaluations through matrix completion to
improve statistical efficiency.
Growing Pains~\cite{habba2026growing} uses fixed
item-response calibration and anchor items to
maintain score comparability across evaluation
periods.
These studies demonstrate the effectiveness of
sparse observations and response modeling.
However, in scaffold evolution, evaluation estimates
also influence archive updates and subsequent
candidate generation.
Therefore, we examine not only estimation
accuracy, but also how shared informative observations
and stable scoring references support reliable
evolutionary decisions.

\textbf{Evaluation Allocation and Candidate Selection.}
Evaluation allocation is closely related to
sequential resource allocation and best-arm
identification.
Successive halving~\cite{jamieson2016sh}
and Hyperband~\cite{li2018hyperband} progressively
concentrate evaluation resources on promising
candidates, while racing-based methods such as
irace~\cite{lopez2016irace} eliminate weaker
configurations through sequential comparisons.
Common random numbers~\cite{glasserman1992crn}
provide a classical motivation for evaluating
alternatives under shared conditions to reduce
the variance of pairwise comparisons.
Recent analyses of code evolution~\cite{simple}
also highlight the importance of reliable candidate
evaluation and selection.
Building on these ideas, our work studies evaluation
allocation across two distinct stages:
providing inexpensive yet comparable feedback
during evolution and reusing accumulated observations
for reliable final scaffold selection.

\section{Problem Formulation}
\label{sec:setting}

\textbf{Agent Scaffold Evaluation.}
Let $c$ denote an executable agent scaffold and $\searchset=\{x_i\}_{i=1}^{N}$ a set of evaluation tasks.
Executing $c$ on $x_i$ produces a score $Y_{ci}^{(r)}\in[0,1]$, where $r$ indexes repeated executions.
Each scaffold--item execution constitutes an evaluation \emph{cell}.
The expected fitness of a scaffold is $F(c)=\frac{1}{N}\sum_{i=1}^{N}\mathbb{E}[Y_{ci}^{(r)}]$.
Evaluating each item once yields the full-set reference score $F_{\mathrm{ref}}(c)=\frac{1}{N}\sum_{i=1}^{N}Y_{ci}^{(1)}$.

\textbf{LLM-Driven Agent Scaffold Evolution.}
At iteration $t$, an evolution framework proposes a candidate scaffold $c_t$ based on its current archive $\archive_t$.
An evaluator selects an item subset $S_t\subseteq\{1,\ldots,N\}$ and returns an estimated fitness $\widehat F_t(c_t)$ using the available observations.
This feedback informs archive updates and subsequent candidate generation.
Evaluation allocation therefore influences the search trajectory and the resulting candidate pool.
After search, a final selector returns a scaffold $c_{\mathrm{pick}}$ from the generated candidates, potentially collecting additional observations.
We distinguish two evaluation stages: \emph{search-time evaluation}, which provides fitness feedback for continued evolution, and \emph{final selection}, which identifies the scaffold to return.
Observations collected during search can be reused during final selection.

\textbf{Cost--Performance Objective.}
Let $\pi$ denote an evaluation allocation policy governing observations collected during both stages and the final scaffold selection.
The total cost $C_{\mathrm{eval}}^{\pi}$ counts executed evaluation cells during search and final selection.
Given an evaluation budget $B$, our objective is
\begin{equation}
    \max_{\pi}
    \;\mathbb{E}\left[F(c_{\mathrm{pick}}^{\pi})\right]
    \quad \text{s.t.} \quad
    \mathbb{E}[C_{\mathrm{eval}}^{\pi}] \leq B,
    \label{eq:objective}
\end{equation}
where $c_{\mathrm{pick}}^{\pi}$ is the scaffold returned under policy $\pi$, and the expectation accounts for randomness in candidate generation, evaluation, and allocation.

\section{Investigation}
\label{sec:findings}

\begin{table}[t]
\caption{Benchmark splits. SEARCH is used by the evaluators during evolution and selection, DEV only  for the audit, and TEST only to score the selected programs.}
\label{tab:splits}
\centering\small
\begin{tabular}{lrrr}
\toprule
Benchmark & SEARCH & DEV & TEST\\
\midrule
HotpotQA & 300 & 150 & 300\\
HoVer & 300 & 150 & 300\\
DROP & 300 & 150 & 300\\
IFBench & 300 & 150 & 294\\
\bottomrule
\end{tabular}
\end{table}

\subsection{Setup}
\label{sec:inv_setup}
We study four benchmarks: HotpotQA~\cite{hotpot}, HoVer~\cite{hover},
DROP~\cite{drop}, and IFBench~\cite{ifbench}.
Each benchmark is divided into SEARCH (300 items), DEV (150 items),
and TEST (300 items, except 294 for IFBench), as shown in Table \ref{tab:splits}.
SEARCH is used for evolution and final selection, DEV for  auxiliary audits, and TEST exclusively for evaluating
the selected scaffolds.
We use two evolution frameworks: ShinkaEvolve~\cite{shinka}
and AdaEvolve~\cite{adaevolve}, the latter implemented using
SkyDiscover~\cite{liu2026skydiscover}.
Each run makes 80 proposals from the same initial program, with three seeds per setting.
We use Qwen3.5-9B~\cite{qwen3.5} as the executor, with temperature zero, thinking disabled, and at most 16,384 output tokens, and DeepSeek-V4-Flash-0731~\cite{xu2026deepseek} as the proposer.
Both models are served locally with vLLM~\cite{kwon2023efficient} on a single node with eight NVIDIA H20 GPUs.

Most analyses in this section replay evaluators on the 24 \emph{Full} runs (four benchmarks, two frameworks, three seeds), in which every candidate is evaluated on all SEARCH items, so the full-SEARCH mean of each candidate is known.
A replay keeps the candidate sequence fixed and varies only which cells the evaluator observes.
We measure two search-time quantities.
\emph{Close-pair agreement} is the share of candidate pairs whose full-SEARCH means differ by at most .03 that the evaluator orders correctly; these are the comparisons that archive updates face.
\emph{Archive regret} is, averaged over the run, the gap between the mean full-SEARCH score of the ten best candidates generated so far and that of the ten candidates the evaluator ranks highest.

\begin{figure}[t]
\centering
\includegraphics[width=0.5\textwidth]{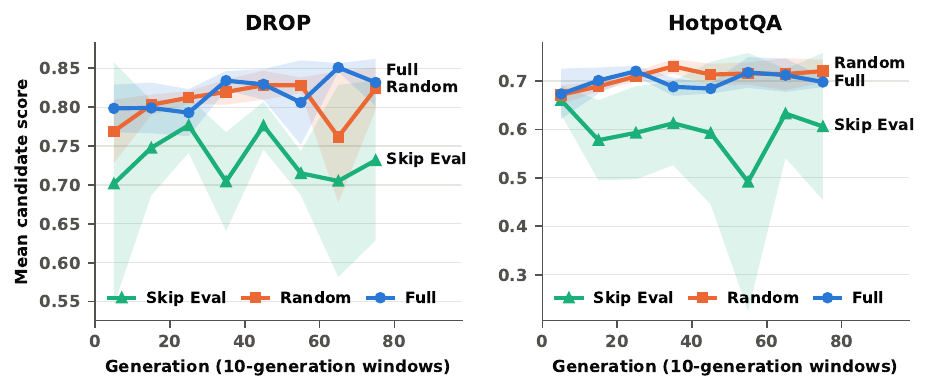}
\caption{Evolution trajectories under different evaluation strategies on DROP and HotpotQA. Curves show mean candidate scores on a shared audit set over successive 10-generation windows.}
\label{fig:inv_search}
\end{figure}

\begin{figure}[t]
\centering
\includegraphics[width=0.5\textwidth]{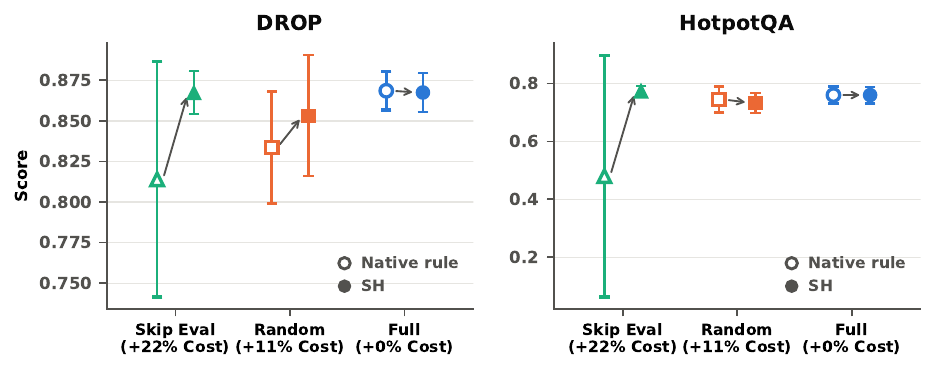}
\caption{Effect of search-time feedback and final selection on DROP and HotpotQA. We compare the native selection rule with successive halving (SH) across three search strategies. Percentages indicate the additional evaluation cost of SH relative to the full-evaluation budget.}
\label{fig:inv_selection}
\end{figure}

\subsection{Search Needs Feedback, but Not Full Evaluation}

Evaluation feedback plays an important role in guiding
evolutionary search.
Figure~\ref{fig:inv_search} compares the evolution
trajectories under three evaluation strategies on
DROP and HotpotQA, using a shared audit set.
Without evaluation feedback, candidate quality remains
relatively low and shows limited improvement.
In contrast, random partial evaluation maintains
candidate quality close to Full, suggesting that
effective evolution does not require evaluating every
candidate on the entire SEARCH set.

Search improvements are also concentrated in early
iterations. 
Although this does not establish the optimal temporal
allocation, it motivates investing in early calibration
and using sparse feedback for subsequent candidates.

\begin{takeawaybox}{Takeaway 1}
Evolutionary search requires evaluation feedback,
but sparse evaluation can sustain candidate improvement
without the cost of full evaluation.
\end{takeawaybox}

\subsection{Comparability Depends on Shared Informative Observations}

We conduct offline replays on 24 Full evolution
trajectories (four benchmarks, two frameworks,
and three seeds), with three replays per trajectory.
Each replay fixes the generated candidate sequence
and varies only the evaluation cells observed by
the evaluator.
The first $W=10$ candidates are fully evaluated,
while subsequent candidates are evaluated on
60 of the 300 SEARCH items.
All policies use the same evaluation budget,
with full-SEARCH scores serving as the reference.

We assess three metrics: Spearman correlation with
full-SEARCH scores, close-pair agreement for candidates
whose reference scores differ by at most .03,
and archive regret, defined as the average fitness gap
between the true and estimated top-10 archives.
These metrics capture both score estimation quality
and the evolutionary decisions supported by
limited feedback.

\textbf{Informativeness and sharing.}
Table~\ref{tab:comparability} shows that shared
informative items yield more reliable close-candidate
comparisons than random or unshared items.
For the unshared informative policy, each candidate
samples 60 items from the 120 most informative ones,
rather than using a common item set.
Despite drawing from informative items, this policy
produces lower close-pair agreement and higher
archive regret.

Pairing provides an explanation for the benefit of
sharing. For $k$ items drawn independently from the
task distribution, the variance of the score difference
between two candidates evaluated on shared items is $
    \operatorname{Var}(\widehat\Delta_{cc'}) =
    \frac{\sigma_c^2+\sigma_{c'}^2-2\sigma_{cc'}}{k},
$
where $\sigma_{cc'}$ denotes their response covariance.
With independently sampled item sets, the covariance
term vanishes.
Because evolved candidates often exhibit correlated
responses, shared items can reduce the variance of
pairwise comparisons.

\textbf{Score comparability.}
Accurate individual rankings do not necessarily
translate into reliable archive decisions.
Although raw means achieve similar Spearman
correlation to model-completed scores, they produce
substantially higher archive regret.
This is because sparse candidates are evaluated
on informative items, whereas warm-up candidates
are evaluated on the full SEARCH set.
The fixed response model places both on a common
full-SEARCH scale, avoiding inconsistent archive
comparisons and repeated rescoring during evolution.

\begin{takeawaybox}{Takeaway 2}
Reliable evolutionary decisions require informative,
shared observations and comparable fitness scores,
rather than accurate individual estimates alone.
\end{takeawaybox}

\begin{table}[t]
\caption{Offline comparison of item policies and fitness
estimators under matched evaluation budgets.
Results are averaged over 24 Full trajectories,
with three replays per trajectory.}
\label{tab:comparability}
\centering
\footnotesize
\setlength{\tabcolsep}{3pt}
\renewcommand{\arraystretch}{1.12}
\begin{adjustbox}{max width=\columnwidth}
\begin{tabular}{@{}llccc@{}}
\toprule
Item policy & Estimator & Spearman
& \shortstack{Close-pair\\agreement}
& \shortstack{Archive\\regret} \\
\midrule
Shared informative   & Model & .862 & .732 & .0043 \\
Shared informative   & Raw   & .865 & .698 & .0102 \\
Unshared informative & Model & .795 & .674 & .0056 \\
Shared random        & Model & .750 & .649 & .0071 \\
Independent random   & Model & .717 & .626 & .0078 \\
\bottomrule
\end{tabular}
\end{adjustbox}
\end{table}


\subsection{Final Selection Needs Confirmation}

Final scaffold selection requires more reliable comparisons
than those needed to guide evolutionary search.
Figure~\ref{fig:inv_selection} compares the native selection
rule with successive halving (SH) on the same candidate
pools under different search-time evaluation strategies.
SH substantially improves final performance when search
feedback is absent, recovering strong candidates that
the native selection rule may overlook.
With random partial evaluation, the gains are smaller
and less consistent, while full evaluation leaves
little room for further improvement.
These results highlight the importance of final
confirmation when search-time evaluation is limited.

\begin{takeawaybox}{Takeaway 3}
Final selection requires reliable comparisons beyond
search-time fitness estimates, particularly when
evaluation feedback is sparse or absent.
\end{takeawaybox}

\section{Method}
\begin{figure*}
    \centering
    \includegraphics[width=0.98\linewidth]{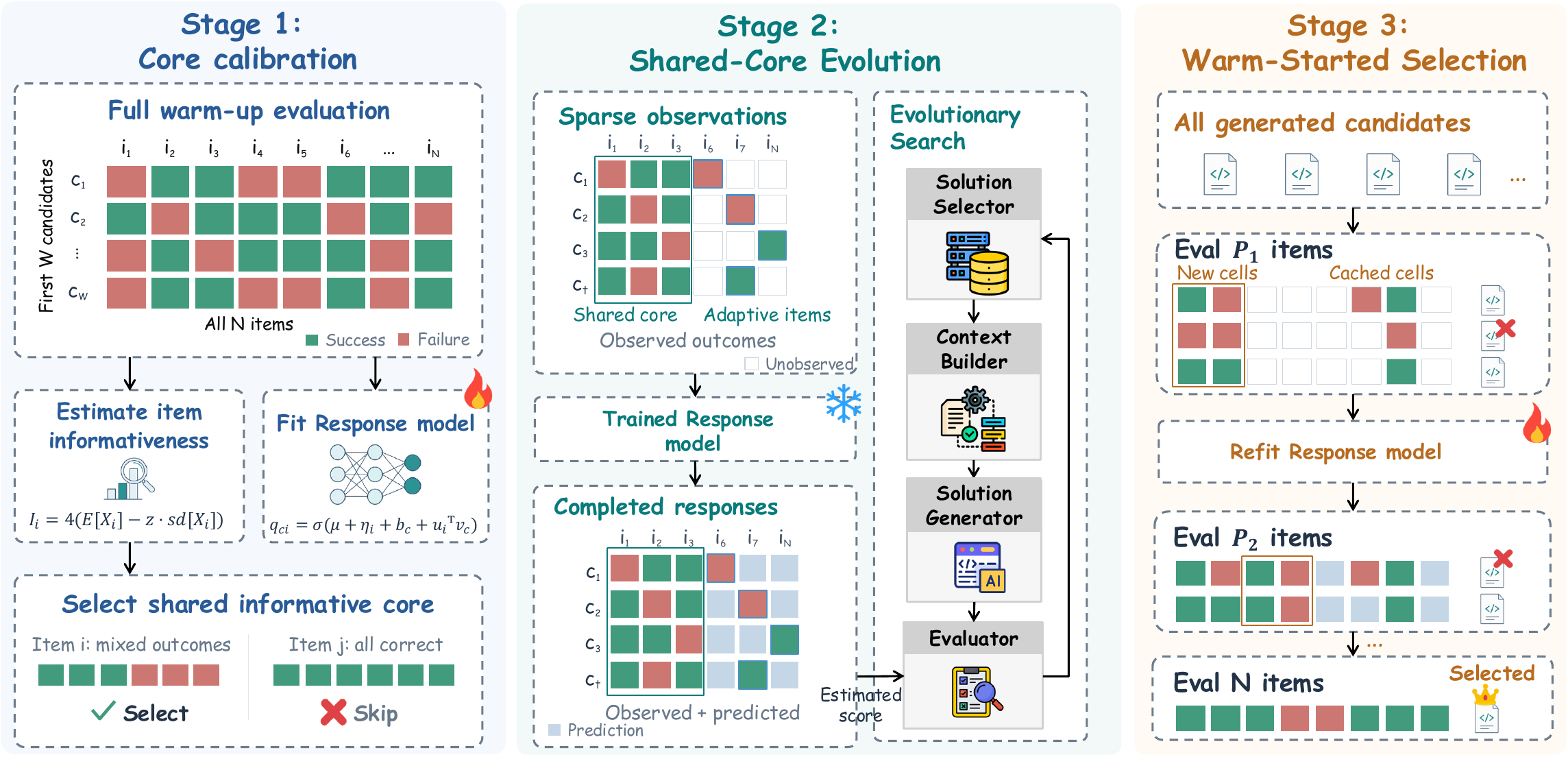}
\caption{Overview of \method{}. The framework consists of three stages: (1) core calibration using fully evaluated warm-up candidates, (2) sparse search-time evaluation with a shared informative core and fixed response model, and (3) warm-started successive halving that reuses accumulated observations for final scaffold selection.}
    \label{fig:main}
\end{figure*}
\label{sec:method}

Figure~\ref{fig:main} illustrates \method{}, an evaluation allocation framework that distributes evaluation effort across three stages: \emph{Core Calibration}, \emph{Shared-Core Evolution}, and \emph{Warm-Started Selection}.
\method{} decouples the two roles of evaluation: search-time feedback only needs to be cheap and comparable enough to guide evolution, whereas the final choice needs precise comparisons among the strongest candidates.
It therefore splits one evaluation budget into a sparse search stage and a separate selection stage that reuses every observation collected during search.
The key idea is to invest a small evaluation budget upfront to establish a reusable reference, use shared informative items to provide comparable feedback throughout evolution, and concentrate additional evaluations on final scaffold selection.
\method{} operates as a drop-in evaluator, leaving the native framework's proposal, archive, and parent-selection policies unchanged.

\subsection{Core Calibration}
\label{sec:warmup}

\method{} begins by fully evaluating the first $W$ valid candidates on all $N$ SEARCH items.
This upfront investment serves two purposes: identifying informative items for subsequent evaluation and calibrating a response model for estimating full-set fitness from sparse observations.
The warm-up candidates receive their directly observed full-set scores, while their item-level outcomes form a dense response matrix for calibration.

\textbf{Identifying informative items.}
We estimate item informativeness from accumulated evaluation outcomes.
Let $a_i$ denote the number of full-credit outcomes and $n_i$ the number of observed outcomes for item $i$.
Using a Beta posterior for its full-credit probability,
\begin{equation}
    p_i \sim \operatorname{Beta}
    (a_i+1,n_i-a_i+1),
\end{equation}
we define a conservative informativeness estimate:
\begin{equation}
    I_i =
    4\left(
        \mathbb{E}[X_i]
        - z\sqrt{\operatorname{Var}(X_i)}
    \right),
    \qquad X_i=p_i(1-p_i).
    \label{eq:info}
\end{equation}
The variance term favors items whose outcomes vary across candidates, while the uncertainty penalty discourages selecting items with insufficient evidence.
Fractional outcomes below full credit count as failures for these acquisition statistics.
The posterior moments are available in closed form.

\textbf{Response model.}
We fit a low-rank logistic response model following a
multidimensional item response theory (MIRT) formulation:
\begin{equation}
    q_{ci} =
    \sigma\!\left(
        \mu + \eta_i + b_c + u_i^\top v_c
    \right),
    \qquad u_i,v_c\in\mathbb{R}^{d},
    \label{eq:model}
\end{equation}
where $\mu$ is a global intercept, $(\eta_i,u_i)$
represent item-specific parameters, and $(b_c,v_c)$
capture candidate-specific response characteristics.
The model is fitted on the fully evaluated warm-up
candidates by minimizing regularized fractional
cross-entropy. Their fitted candidate parameters
also define a Gaussian prior for subsequent candidates.

During evolution, the global intercept, item parameters,
and candidate prior remain fixed.
For each new candidate, only $(b_c,v_c)$ are inferred
from its sparse observations by MAP estimation under
this prior.
This maintains a common scoring reference across
candidates without repeatedly refitting the model
or rescoring archived candidates.

\subsection{Shared-Core Evolution}
\label{sec:sparse}

After calibration, each new candidate is evaluated on a small subset of $k$ SEARCH items ($k=60$, i.e., 20\% of $N=300$).
Rather than independently selecting items for each candidate, \method{} combines a fixed informative core with a small adaptive component.

\textbf{Shared informative core.}
Using the warm-up informativeness estimates, we select the top $k_0$ items to form a fixed core $S_0$.
Every subsequent candidate is evaluated on this same core, creating a common basis for cross-candidate comparisons.
This design follows the intuition of common random numbers (CRN): exposing candidates to shared evaluation conditions helps reduce variation arising from differences in the evaluated item sets.

To retain limited adaptability, the remaining $k-k_0$ items are candidate-specific.
For each candidate $c$, we sample a set $A_c$ from outside the core, using Gumbel-top-$k$ sampling with logits $I_i/\tau$ computed from the current informativeness scores.
With a small temperature, this amounts to taking the currently most informative items with random tie-breaking.
The resulting evaluation set is
\begin{equation}
    S_c=S_0\cup A_c,
    \qquad
    |S_0|=k_0,\quad |S_c|=k.
    \label{eq:allocation}
\end{equation}
The shared core remains unchanged, whereas the acquisition statistics are updated after each candidate using newly observed outcomes.
This allows the adaptive component to respond to changes in the evolving candidate population without sacrificing the common comparison basis.

Concretely, for two candidates scored on the same $k_0$ items, the variance of their score difference is
$\operatorname{Var}(\bar Y_a-\bar Y_b)=\big(\sigma_a^2+\sigma_b^2-2\operatorname{Cov}(Y_a,Y_b)\big)/k_0$,
whereas independently drawn item sets remove the covariance term.
Because evolved scaffolds are strongly correlated---competitive candidates agree on most items---a shared core makes pairwise comparisons less noisy at the same budget.

\textbf{Comparable fitness feedback.}
After evaluating candidate $c$ on $S_c$, we infer its candidate-specific response parameters under the fixed warm-up model.
The fitted model then predicts responses on unobserved items, producing a completed estimate of full-set fitness:
\begin{equation}
    \widehat F(c)=
    \frac{1}{N}
    \left(
        \sum_{i\in S_c}Y_{ci}
        +
        \sum_{i\notin S_c}q_{ci}
    \right).
    \label{eq:fitness}
\end{equation}
Here, $Y_{ci}$ denotes the observed outcome and $q_{ci}$ the model prediction using the inferred parameters of candidate $c$.

Combining observed and predicted outcomes places sparse evaluations on the full-SEARCH fitness scale expected by the native evolution framework, which also contains the fully evaluated warm-up candidates.
Keeping the response model fixed also avoids retrospectively changing the scoring reference for archived candidates.
The resulting fitness estimates are returned directly to the framework for its existing archive updates and subsequent proposals.
All observed outcomes are cached for reuse during final selection.

\subsection{Warm-Started Selection}
\label{sec:selector}

Search-time evaluation aims to provide inexpensive feedback for continued evolution, whereas final selection requires distinguishing the strongest candidates.
After search, \method{} therefore runs successive halving (SH) over all valid candidates, using additional observed outcomes rather than relying on search-time fitness estimates.
Candidates with identical code (the same abstract syntax tree after removing docstrings) are merged first.

We construct a shared random permutation of the SEARCH items and define increasingly large cumulative prefixes
$P_1\subset P_2\subset\cdots\subset P_R$ with $P_R=\searchset$.

In each round, surviving candidates are evaluated on the current prefix, and the top candidates are retained according to their round scores.
The evaluation budget thus becomes increasingly concentrated on competitive candidates.

\textbf{Warm-starting from search observations.}
Rather than discarding evaluations collected during evolution, we reuse all available outcomes.
Let $O_c$ denote the items already observed for candidate $c$ during search, and let $U_c^r=O_c\cup P_r$.
Only the missing observations in $U_c^r$ are executed; cached outcomes are reused without additional cost.
In the first round, the score is the observed mean over this union:
\begin{equation}
    \widetilde F_1(c)=
    \frac{1}{|U_c^1|}
    \sum_{i\in U_c^1}Y_{ci}.
    \label{eq:union}
\end{equation}

\begin{table*}[!t]
\caption{
Test performance of the program selected under each evaluation strategy.
For each benchmark, Score reports the TEST mean over three seeds, with standard deviation shown as a subscript.
Cost (\%) is the evaluation spent during search and final selection, as a share of evaluating every proposal on all 300 search items ($80\times300$ cells for ShinkaEvolve, $81\times300$ for AdaEvolve); programs that fail to load or error on every evaluated item cost nothing.
Average reports the unweighted mean over the four benchmarks.
The highest Score per benchmark and on average within each framework is bold.
}
\label{tab:main}
\centering
\small
\setlength{\tabcolsep}{3.5pt}
\begin{tabular}{llcccccccccc}
\toprule
\multirow{2}{*}{Framework}
& \multirow{2}{*}{Method}
& \multicolumn{2}{c}{DROP}
& \multicolumn{2}{c}{HotpotQA}
& \multicolumn{2}{c}{HoVer}
& \multicolumn{2}{c}{IFBench}
& \multicolumn{2}{c}{Average} \\
\cmidrule(lr){3-4}
\cmidrule(lr){5-6}
\cmidrule(lr){7-8}
\cmidrule(lr){9-10}
\cmidrule(lr){11-12}
& & Score $\uparrow$ & Cost $\downarrow$ & Score $\uparrow$ & Cost $\downarrow$
  & Score $\uparrow$ & Cost $\downarrow$ & Score $\uparrow$ & Cost $\downarrow$ & Score $\uparrow$ & Cost $\downarrow$ \\
\midrule

\multirow{6}{*}{ShinkaEvolve}
& Skip Eval
& $0.7674_{\pm 0.1046}$ & 0.0
& $0.4633_{\pm 0.4014}$ & 0.0
& $0.3933_{\pm 0.0961}$ & 0.0
& $0.3917_{\pm 0.0741}$ & 0.0
& $0.5039$ & 0.0 \\

& Successive Halving
& $0.8738_{\pm 0.0248}$ & 22.1
& $0.7089_{\pm 0.0139}$ & 22.0
& $0.6844_{\pm 0.0948}$ & 22.5
& $0.4433_{\pm 0.0098}$ & 21.2
& $0.6776$ & 22.0 \\

& Random
& $0.8629_{\pm 0.0083}$ & 42.5
& $0.7244_{\pm 0.0158}$ & 45.6
& $0.7156_{\pm 0.0443}$ & 43.8
& $0.4603_{\pm 0.0332}$ & 37.7
& $0.6908$ & 42.4 \\

& Cascade
& $0.8643_{\pm 0.0139}$ & 39.6
& $0.7178_{\pm 0.0135}$ & 53.3
& $\mathbf{0.7289}_{\pm 0.0375}$ & 54.8
& $0.4683_{\pm 0.0214}$ & 38.5
& $0.6948$ & 46.6 \\

& Full
& $0.8832_{\pm 0.0155}$ & 77.1
& $0.7133_{\pm 0.0067}$ & 92.5
& $0.7267_{\pm 0.0176}$ & 92.5
& $0.4660_{\pm 0.0290}$ & 75.4
& $0.6973$ & 84.4 \\

\rowcolor{lightcyan} \cellcolor{white}& \method{}
& $\mathbf{0.8907}_{\pm 0.0164}$ & 43.3
& $\mathbf{0.7411}_{\pm 0.0234}$ & 46.0
& $0.7278_{\pm 0.0847}$ & 46.9
& $\mathbf{0.4739}_{\pm 0.0246}$ & 39.2
& $\mathbf{0.7084}$ & 43.8 \\

\midrule

\multirow{6}{*}{AdaEvolve}
& Skip Eval
& $0.7720_{\pm 0.0833}$ & 0.0
& $0.6678_{\pm 0.0287}$ & 0.0
& $0.4933_{\pm 0.1389}$ & 0.0
& $0.4303_{\pm 0.0398}$ & 0.0
& $0.5908$ & 0.0 \\

& Successive Halving
& $0.8559_{\pm 0.0480}$ & 22.8
& $0.6922_{\pm 0.0402}$ & 23.2
& $0.5622_{\pm 0.1155}$ & 23.1
& $0.4478_{\pm 0.0354}$ & 23.1
& $0.6395$ & 23.1 \\

& Random
& $0.8759_{\pm 0.0098}$ & 49.2
& $0.7167_{\pm 0.0384}$ & 49.8
& $0.7178_{\pm 0.0877}$ & 49.6
& $0.4473_{\pm 0.0173}$ & 48.6
& $0.6894$ & 49.3 \\

& Cascade
& $0.8204_{\pm 0.0398}$ & 58.6
& ${0.7378}_{\pm 0.0356}$ & 53.8
& $\mathbf{0.7922}_{\pm 0.1091}$ & 60.3
& $0.4518_{\pm 0.0221}$ & 51.4
& $0.7006$ & 56.0 \\

& Full
& $0.8840_{\pm 0.0155}$ & 100.0
& $0.7267_{\pm 0.0296}$ & 98.8
& $0.7733_{\pm 0.0899}$ & 100.0
& $0.4524_{\pm 0.0123}$ & 97.9
& $0.7091$ & 99.2 \\

\rowcolor{lightcyan} \cellcolor{white}& \method{}
& $\mathbf{0.8881}_{\pm 0.0087}$ & 46.3
& $\mathbf{0.7456}_{\pm 0.0310}$ & 46.9
& $0.7844_{\pm 0.0619}$ & 47.3
& $\mathbf{0.4722}_{\pm 0.0513}$ & 45.3
& $\mathbf{0.7226}$ & 46.4 \\

\bottomrule
\end{tabular}
\end{table*}

\textbf{Regression-adjusted scores.}
Because $O_c$ contains informative and partly candidate-specific items, union means of different candidates are computed on item sets of different difficulty.
From the second round on, \method{} corrects this with a regression (difference) estimator~\cite{cochran1977sampling,angelopoulos2026ppi}:
\begin{equation}
    \widetilde F_r(c)=
    \frac{1}{|U_c^r|}\sum_{i\in U_c^r}Y_{ci}
    +\lambda_r\left(
    \frac{1}{N}\sum_{i=1}^{N}\tilde q_{ci}
    -\frac{1}{|U_c^r|}\sum_{i\in U_c^r}\tilde q_{ci}
    \right),
    \label{eq:regest}
\end{equation}
where $\tilde q_{ci}$ is the prediction of a refitted response model.
The second term adds the predicted difference between the full SEARCH set and the observed items, so $\lambda_r=0$ recovers the union mean and $\lambda_r=1$ the classical difference estimator.
We estimate $\lambda_r$ from the data as the pooled within-candidate regression slope of observed outcomes on predictions over the surviving candidates $\mathcal C_r$,
\begin{equation}
    \lambda_r=
    \frac{\sum_{c\in\mathcal C_r}\sum_{i\in U_c^r}(Y_{ci}-\bar Y_c)(\tilde q_{ci}-\bar{\tilde q}_c)}
         {\sum_{c\in\mathcal C_r}\sum_{i\in U_c^r}(\tilde q_{ci}-\bar{\tilde q}_c)^2},
    \label{eq:lambda}
\end{equation}
with means taken over $U_c^r$ and the result clipped to $[0,1.5]$.
This slope approximates the variance-minimizing coefficient of the difference estimator and automatically shrinks the correction when predictions track the observed outcomes poorly.

\textbf{Cross-fitted refit.}
After the first round, we refit the response model on all search-time and round-1 observations.
To avoid fitting the item parameters that correct a candidate on that candidate's own observations, we use two-fold cross-fitting~\cite{chernozhukov2018double}: the sparse (non-warm-up) candidates are randomly split into two folds, and each fold is scored by a model fitted on the warm-up candidates and the other fold.
Warm-up candidates are fully observed, so their correction term is zero.
Likewise, once the last round observes all $N$ items, $U_c^R=\searchset$ and Eq.~\eqref{eq:regest} reduces to the observed full-SEARCH mean, so the final pick is always decided by observed outcomes.

\textbf{Evaluation cost.}
For $M$ valid candidates, the total evaluator cost in executed cells is
\begin{equation}
    C_{\mathrm{eval}}
    =
    WN+(M-W)k+C_{\mathrm{SH,new}},
    \label{eq:method_cost}
\end{equation}
where the first two terms account for calibration and sparse search, and $C_{\mathrm{SH,new}}$ counts only additional executions required by successive halving.
With 80 valid proposals, the search stage costs $(WN+70k)/(80N)=30\%$ of full evaluation.
Thus, \method{} reuses evaluation effort across stages: dense observations establish the scoring reference, shared sparse observations guide evolution, and cached evaluations reduce the incremental cost of final confirmation.

\begin{table*}[t]
\caption{Ablation of \method{} on ShinkaEvolve. For each benchmark, we report the score over three runs, with standard deviation shown as a subscript. Each variant removes one component and keeps the rest unchanged. $\Delta$ is the change in the four-benchmark average relative to \method{}. Cost (\%) is the evaluation spent during search and final
selection. Best value per column is bold.}
\label{tab:ablation_live}
\centering
\small
\setlength{\tabcolsep}{3pt}
\begin{adjustbox}{max width=1.8\columnwidth}
\begin{tabular}{lccccccc}
\toprule
Variant & DROP & HotpotQA & HoVer & IFBench & Avg. & $\Delta$ & Cost \\
\midrule
\rowcolor{lightcyan} \method{}
& $0.8907_{\pm 0.0164}$ & $\mathbf{0.7411}_{\pm 0.0234}$ & $\mathbf{0.7278}_{\pm 0.0847}$ & $0.4739_{\pm 0.0246}$
& $\mathbf{0.7084}$ & -- & 43.8\% \\
\quad \textit{w/o} selection
& $0.8755_{\pm 0.0157}$ & $0.7278_{\pm 0.0139}$ & $0.6922_{\pm 0.0683}$ & $0.4688_{\pm 0.0363}$
& $0.6911$ & $-0.0173$ & 27.4\% \\
\quad \textit{w/o} response model
& $\mathbf{0.8946}_{\pm 0.0113}$ & $0.7222_{\pm 0.0139}$ & $0.6800_{\pm 0.0361}$ & $\mathbf{0.4892}_{\pm 0.0513}$
& $0.6965$ & $-0.0119$ & 44.2\% \\
\quad \textit{w/o} informative items
& $0.8866_{\pm 0.0028}$ & $0.6967_{\pm 0.0260}$ & $0.6689_{\pm 0.0369}$ & $0.4649_{\pm 0.0326}$
& $0.6793$ & $-0.0291$ & 43.4\% \\
\bottomrule
\end{tabular}
\end{adjustbox}
\end{table*}

\section{Experiments}
\label{sec:experiments}
\subsection{Setup}

\textbf{Baselines.}
We compare \method against five evaluation strategies with different budget allocations.
\textbf{Skip Eval} performs no evaluation during either search or final selection, and select the last valid evolved program.
\textbf{Successive Halving (SH)} \cite{jamieson2016sh} provides no evaluation feedback during search and applies successive halving to the generated candidate pool for final selection.
\textbf{Random} evaluates each candidate on a randomly sampled 50\% subset of SEARCH items.
\textbf{Cascade} \cite{novikov2025alphaevolve,gepa} first evaluates candidates on 20\% of SEARCH items and promotes candidates meeting a top-25\% threshold to full evaluation, while allocating no additional evaluations to the remaining candidates. 
\textbf{Full} evaluates every candidate on all SEARCH items.

\textbf{Implementation Details.}
We use the same benchmarks, evolution frameworks, and model configurations as in Section~4.1.
Unless otherwise specified, \method allocates 30\% of the full-evaluation budget to search, with the first $W=10$ valid candidates fully evaluated and each subsequent candidate evaluated on $k=60$ items (20\% of SEARCH).
The shared informative core contains $k_0=54$ items (90\% of $k$), and the Gumbel sampling temperature is $\tau=0.001$.
For final selection, we use $R=4$ successive-halving rounds with cumulative item prefixes of 30, 60, 120, and 300, retaining 30, 15, 8, and 1 candidates, respectively.
For the response model, we set the latent dimension to $d=2$, the regularization coefficient to $\rho=1$, and the informativeness uncertainty penalty to $z=2$.


\subsection{Main Results}

Table~\ref{tab:main} compares the TEST performance and evaluation cost of different strategies.
Across both evolution frameworks and all four benchmarks, \method{} matches or slightly exceeds the mean TEST score of Full while using approximately half as many evaluation cells.
It also outperforms Random and Cascade in cross-benchmark average performance at comparable or lower evaluation cost.
These results demonstrate that allocating evaluations selectively can provide a better cost--performance tradeoff than uniformly evaluating candidates.
The evaluation costs differ between the two frameworks partly due to their handling of invalid proposals.
AdaEvolve retries program generation when a proposal is invalid, whereas ShinkaEvolve discards invalid proposals without replacement.
Since invalid programs incur no evaluation cost, ShinkaEvolve evaluates fewer valid candidates, resulting in lower realized evaluation costs, particularly under Full evaluation.

The no-evaluation baselines further highlight the complementary roles of search feedback and final selection.
Applying SH to candidates generated without evaluation feedback substantially improves final performance, but still falls short of \method{}.
This suggests that final selection can recover promising candidates even from unguided search, while informative feedback during evolution provides additional benefits.
These findings support combining sparse search-time evaluation with targeted final confirmation.

\subsection{Ablation Study}
\label{sec:ablation}

We ablate three components of \method{} on ShinkaEvolve across all four benchmarks.
\emph{W/o selection} skips final confirmation and selects the candidate with the highest refitted response-model estimate from the same search runs.
\emph{W/o response model} uses raw observed means for search feedback and raw union means for final selection.
\emph{W/o informative items} replaces informative item selection with random sampling while preserving the shared core and evaluation budget.
The latter two variants require new evolution runs because they change the search-time feedback.

Table~\ref{tab:ablation_live} shows that removing final confirmation consistently reduces TEST performance, highlighting the importance of additional evaluation for selecting among competitive candidates.
Replacing informative items with random ones produces the largest average degradation, suggesting that sharing evaluation items alone is insufficient without considering their informativeness.
Removing the response model has a smaller and less consistent effect, as the shared core already provides substantial comparability among candidates.
The response model therefore primarily serves to calibrate sparse feedback onto a common scoring scale.
Overall, these results identify informative item selection and final confirmation as the main contributors to \method{}.


\subsection{Analysis}

\begin{figure}[t]
    \centering
    \includegraphics[width=1\linewidth]{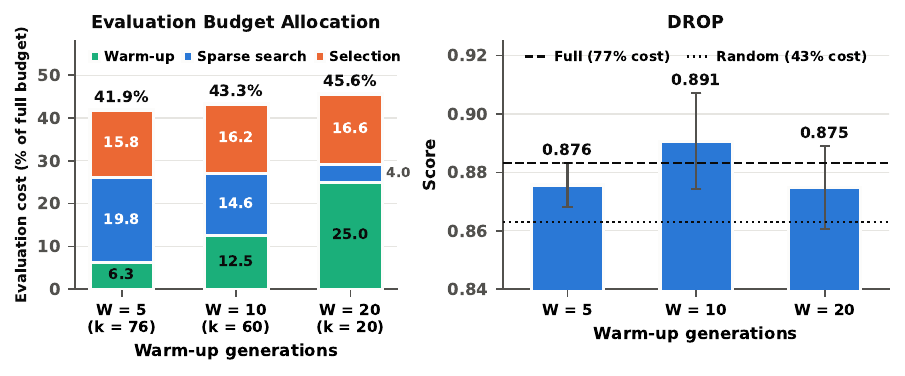}
    \caption{Effect of warm-up length on DROP with ShinkaEvolve under a fixed search-stage budget. Left: evaluation cost breakdown for different numbers of fully evaluated warm-up candidates. Right: corresponding performance, with Full and Random shown as references.}
    \label{fig:warm}
\end{figure}

\textbf{Budget Allocation in Warm-up Stage.}
Figure~\ref{fig:warm} examines the effect of warm-up length $W$ on DROP with ShinkaEvolve under a fixed search-stage budget.
A longer warm-up provides more observations for calibration but leaves fewer evaluation items for subsequent candidates.
The default $W=10$ achieves the highest mean  performance, although the differences are within seed-to-seed variation.
All three configurations outperform Random at comparable evaluation cost.
Offline replays further show that an excessively long warm-up reduces search-time ranking accuracy as later evaluations become too sparse.
These results highlight the trade-off between upfront calibration and informative feedback during subsequent evolution.

\begin{figure}
    \centering
    \includegraphics[width=\linewidth]{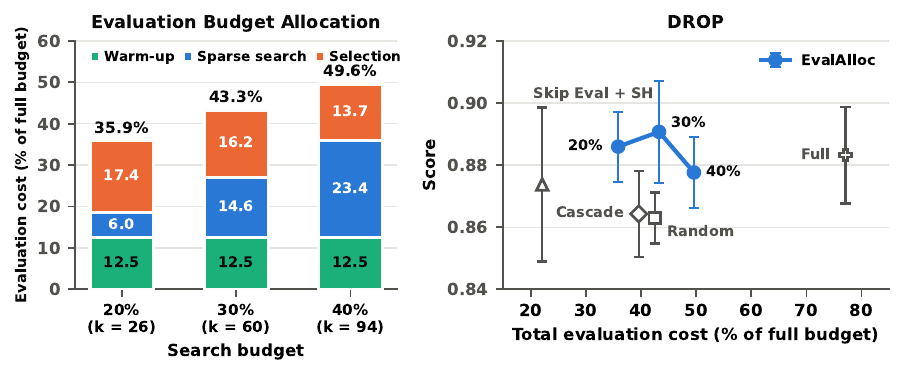}
\caption{Effect of search-time evaluation budget on DROP with ShinkaEvolve. Left: evaluation cost breakdown across warm-up, sparse search, and final selection. Right: Performance versus total evaluation cost under different search budgets.}
    \label{fig:search_budget}
\end{figure}

\textbf{Budget Allocation in Search Stage.}
Figure~\ref{fig:search_budget} examines the effect of search-time evaluation budget on DROP with ShinkaEvolve.
We fix the warm-up at ten fully evaluated candidates and vary the target search budget from 20\% to 40\%.
The default 30\% allocation achieves the highest observed mean TEST performance, while the 20\% configuration offers comparable performance at lower cost.
Increasing the search-time budget does not consistently improve final performance in this setting, suggesting that sparse feedback can be sufficient to guide productive evolution.

\begin{figure}
    \centering
    \vspace{-0.1in}\includegraphics[width=\linewidth]{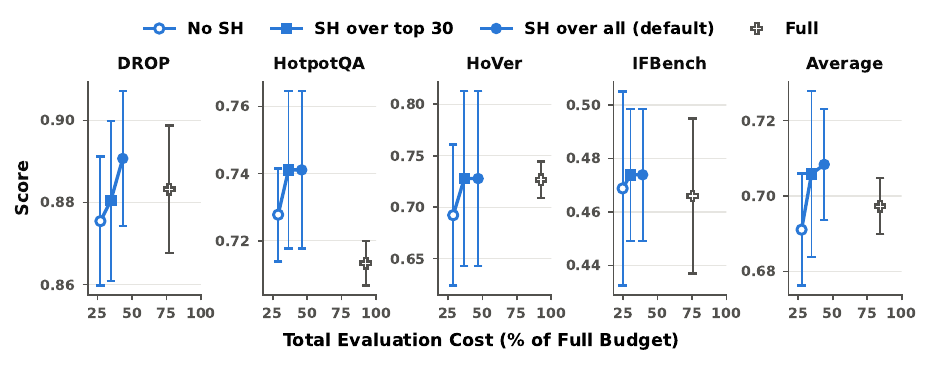}
    \caption{Effect of final selection strategies on TEST performance and evaluation cost across four benchmarks and their average. We compare selection without SH, SH over the top 30 candidates, and SH over all candidates, using the same search runs. Full evaluation is shown as a reference.}
    \label{fig:confirm_budget}
\end{figure}

\textbf{Budget Allocation in Selection Stage.}
Figure~\ref{fig:confirm_budget} examines the effect of final confirmation budget while reusing the same search runs.
Compared with direct selection based on search-stage estimates, successive halving (SH) improves TEST performance across all four benchmarks.
Applying SH to the top 30 candidates captures most of the improvement at lower cost, while considering all candidates provides a modest additional gain.
These results highlight the value of targeted final confirmation and the trade-off between selection cost and final scaffold quality.

\begin{table}[t]
\caption{
Evaluation-cell and executor-token (prompt and completion) costs of search and final selection, relative to Full (100\%).
Each entry averages the per-benchmark shares over the four benchmarks; programs that fail to load or error on every evaluated item cost nothing.
}
\label{tab:token_cost}
\centering
\small
\setlength{\tabcolsep}{4pt}
\begin{tabular}{lrrrr}
\toprule
\multirow{2}{*}{Method}
& \multicolumn{2}{c}{ShinkaEvolve}
& \multicolumn{2}{c}{AdaEvolve} \\
\cmidrule(lr){2-3}
\cmidrule(lr){4-5}
& Cells & Tokens & Cells & Tokens \\
\midrule
Full
& 100.00\% & 100.00\% & 100.00\% & 100.00\% \\
Successive Halving
& 26.26\% & 27.50\% & 23.26\% & 15.10\% \\
Random 
& 50.44\% & 55.18\% & 49.69\% & 43.66\% \\
Cascade
& 54.83\% & 66.32\% & 56.47\% & 54.50\% \\
\rowcolor{lightcyan}
\method{}
& 52.13\% & 64.76\% & 46.83\% & 45.47\% \\
\bottomrule
\end{tabular}
\end{table}

\textbf{Evaluation cost.}
Table~\ref{tab:token_cost} compares evaluation-cell
and executor-token costs relative to Full.
\method{} reduces token consumption across all
benchmark--framework settings, although token savings
are not always proportional to cell savings.
On ShinkaEvolve, it uses 52.13\% of Full's cells
but 64.76\% of its tokens, whereas its token share
on AdaEvolve is 45.47\%.
The latter average is influenced by an unusually
token-intensive HoVer Full run.

To examine this discrepancy without confounding
differences in search trajectories, we analyze the
fully evaluated candidate pools from the Full runs.
Higher-scoring candidates tend to consume more tokens
per item, with average within-run Spearman correlations
of 0.25 and 0.36 for ShinkaEvolve and AdaEvolve,
respectively.
Likewise, the 60 items with the highest response
variability consume 1.08 and 1.13 times the average
tokens per item.
These patterns suggest that concentrating evaluations
on informative items and competitive candidates can
favor more expensive execution cells.
Thus, evaluation-cell counts measure the number of
executions, while token counts additionally reflect
heterogeneity in execution cost.

\begin{figure}
    \centering
    \includegraphics[width=1\linewidth]{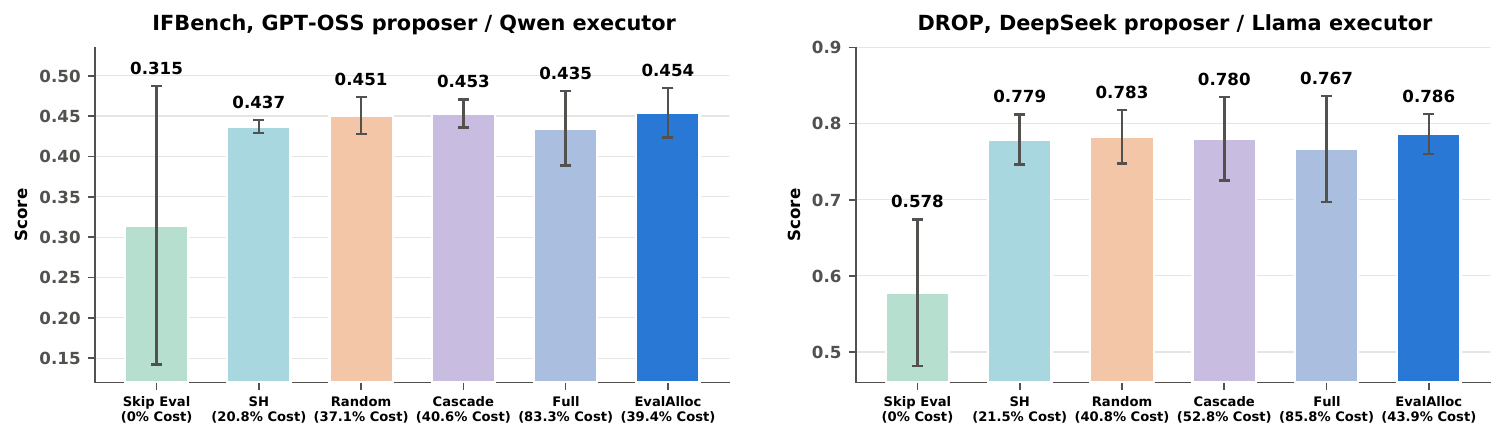}
    \caption{Performance with alternative proposer
and executor models on ShinkaEvolve.
Left: IFBench with a gpt-oss-120b proposer.
Right: DROP with a Llama-3.1-8B-Instruct executor.
Bars show the mean over three seeds, with standard
deviation as error bars. Percentages indicate
evaluation cost relative to the nominal full budget.}
    \label{fig:model_transfer}
\end{figure}

\textbf{Other Proposers and Executor.}
\label{app:proposer}
We further evaluate alternative model configurations
on ShinkaEvolve in Figure~\ref{fig:model_transfer}.
We replace the proposer with gpt-oss-120b \cite{agarwal2025gpt} on IFBench,
or the executor with Llama-3.1-8B-Instruct \cite{grattafiori2024llama} on DROP.
In both settings, \method{} achieves competitive
TEST performance with substantially fewer evaluation
cells than Full.
The relative performance of evaluation strategies
varies across model configurations, suggesting that
the benefits of evaluation allocation can depend
on the underlying models.

\section{Conclusion}

Evaluation influences both evolutionary search and
final scaffold selection.
Our investigation shows that sparse informative
feedback can guide evolution, while shared evaluation
items and a stable scoring reference support
comparable decisions across candidates.
Motivated by these findings, we propose \method{},
which combines sparse search-time feedback with
warm-started final confirmation.
Across four benchmarks and two evolution frameworks,
\method{} achieves competitive performance
while using approximately half as many evaluation
cells as Full.
These results highlight the importance of allocating
evaluation effort according to its distinct roles
in search and final selection.

\bibliographystyle{plainnat}

\bibliography{ref}

\providecommand{\tbd}{\textcolor{red}{TBD}}

\providecommand{\tbd}{\textcolor{red}{TBD}}

\appendix

\end{document}